# LIP: Lightweight Intelligent Preprocessor for meaningful text-to-speech

Harshvardhan Anand †, Nansi Begam †, Richa Verma †, Sourav Ghosh §, Harichandana B.S.S §, Sumit Kumar §

† *Dept. of Electronics and communication, Sikkim Manipal Institute of Technology*, India
§ *Samsung R&D Institute Bangalore*, India

hharshvardhanaanand@gmail.com, nansibegum@gmail.com, richaverma6561@gmail.com,
sourav.ghosh@samsung.com, hari.ss@samsung.com, sumit.kr@samsung.com

***Abstract*—Existing Text-to-Speech (TTS) systems need to read messages from the email which may have Personal Identifiable Information (PII) to text messages that can have a streak of emojis and punctuation. 92% of the world's online population use emoji[1] with more than 10 billion emojis sent everyday[2]. Lack of preprocessor leads to messages being read as-is including punctuation and infographics like emoticons. This problem worsens if there is a continuous sequence of punctuation/emojis that are quite common in real-world communications like messaging, Social Networking Site (SNS) interactions etc. [1]. In this work, we aim to introduce a lightweight intelligent preprocessor (LIP) that can enhance the readability of a message before being passed downstream to existing TTS systems. We propose multiple sub-modules including: expanding contraction, censoring swear words, and masking of PII, as part of our preprocessor to enhance the readability of text. With a memory footprint of only 3.55 MB and inference time of 4 ms for up to 50-character text, our solution is suitable for real-time deployment. This work being the first of its kind, we try to benchmark with an open independent survey, the result of which shows 76.5% preference towards LIP enabled TTS engine as compared to standard TTS.**

## I. Introduction

With the rapid advance in deep learning and artificial intelligence, Text-to-Speech (TTS) technology being an important task in speech processing, has received widespread attention in recent years. Modern TTS systems have come a long way in using neural networks leading to a dramatic improvement in performance, especially through effectively learning various speaking styles thus generating high-quality outputs mimicking natural human voice.

According to a study [2], the TTS market was valued at USD 2.0 billion in 2020 and is estimated to reach USD 5.0 billion by 2026. The rising demand for handheld devices, increased government spending on education for differently-abled, the dependence of the growing elderly population on technology, and the rising number of people with different learning disabilities or learning styles are factors driving the growth of the text-to-speech market. There is increased attention to developing interfaces and software specifically targeted at enabling people with disabilities as they comprise around 15% of the world population [3] [4]. In another report by Microsoft [5] where they made an effort to Understand smartphone usage patterns and practices of people with visual impairments, it is found that owing to the visually demanding nature of smartphones, TTS is one of the features that is frequently relied upon by them. Thus, even though ample effort is being put in to introduce improved software input methods like in the work of Ghosh et al. [6], similar research investment is lacking to enhance output systems like TTS.

Existing TTS system reads the emojis as general text, resulting in a poor user experience. Let's consider an example:

*"The show was awesome 🤣🤣I liked it"*

is read as:

*"The show was awesome rolling on the floor laughing rolling on the floor laughing I liked it."*

Now the user is unable to comprehend the message but our system retains the content and properly describes the emoji.

*"The show was awesome I liked it with rolling on the floor laughing emoji."*

Moreover, messages may also include emojis that provide information. for example –

" 2️⃣4️⃣ *HOURS 'till our schedule drops!*

Our system can comprehend a message and retain the position of informational emoji (such as 2 and 4), as well as mask PII such as emails, phone numbers, and so on. Text normalisation feature corrects common misspellings and contractions such as *"msg"* or *"messg"* to *"message"*.

A text-to-speech technology is available for every digital device to assist people who have difficulty reading on-screen text, but what good is it if it can't generate user-friendly output? For example, TTS will read the message with a phone number *"9312586790"* as *"nine billion, three hundred and twelve million, five hundred and eighty-six thousand, seven hundred and ninety"*.The idea behind our TTS pre-processor is to teach the system how we humans comprehend any message, which makes it incredibly user-friendly; the output of the above message is *"nine three one two five eight six seven nine zero"*.

[1] https://home.unicode.org/emoji/emoji-frequency/

[2] https://www.brandwatch.com/blog/6-facts-about-emojis-found-using-new-analysis/

Our contributions in this paper can be summarized as:

1) Pre-process messages that have emoji and punctuation.
2) Pre-process personally Identifiable information like phone number, email, aadhar card number etc.
3) Better presentation of URLs.
4) Better presentation of phone numbers.
5) Censoring swear words
6) Handling message contraction and misspells like *"msg"* or *"meeesssagee"* is converted to *"message"* to pronounce it properly.

## II. Related Work

Cohn et al. [7] conducted experiments to understand how do emojis interact with the grammar of written text and found that emojis lack grammatical structure on their own whereas they are substituted for nouns and adjectives, while also typically conveying non-redundant information to the sentences. Berengueres et al. [8] on comparing differences in emoji sentiment perception between readers and writer found that there is an 82% agreement in how emoji sentiment is perceived by reader and writer but the disagreement concentrates on negative emoji with 26% worse than perceived by readers. The placement of an emoji in its textual context can determine its role as an amplifier or modifier of the emotional range of a message. Research by Novak et al. [9] suggests that the typical emoticon user employs icons sparingly and preferably at the end of a sentence; emoji, in contrast, are more likely to be grouped and their placement determined by emotional content. The most beneficial for the speech synthesis system is the situation in which there is an unambiguous relationship between spelling and pronunciation. But in real text, there are many unusual words: numbers, digit sequences, acronyms, abbreviations, dates. The main problem of the text normalization module is converting non-standard words into regular words [10]. Hassan et al. [11] have developed a text normalization system based on unsupervised learning of the normalization equivalences from the unlabeled text. It uses random walks on a contextual similarity bipartite graph constructed from n-gram sequences on a large unlabeled text corpus. Their algorithm has achieved an F-Measure of 70.05% but it is quite heavy to deploy on a mobile device. Most of the text was never written with the intention that it will be read aloud, and because of this, faithful reading of a text can lead to situations where the reader is speaking something that the listener cannot understand [10] for example link or national ID or any other PII. Various ideas have been implemented to protect personally identifiable information using deep learning techniques like Facebook's RoBERTa that presents a higher F1-score but at the cost of speed, to DistilBERT that compromises between speed and accuracy [12] and by comparing the F1-score of NLTK, Stanford CoreNLP and spaCy is approximately 0.67, 0.84 and 0.86, respectively [13].

Our proposed solution outperforms existing systems both in terms of memory footprint and inference time that is essential for real-time on-device performance. This is achieved by leveraging preloaded models that rely on regular expressions and dictionaries instead of heavy ML models or embeddings with a huge number of parameters. Furthermore, our pipeline is designed to be customizable, modular and plug-and-play, with a scope of personalization and extension.

## III. Methodology

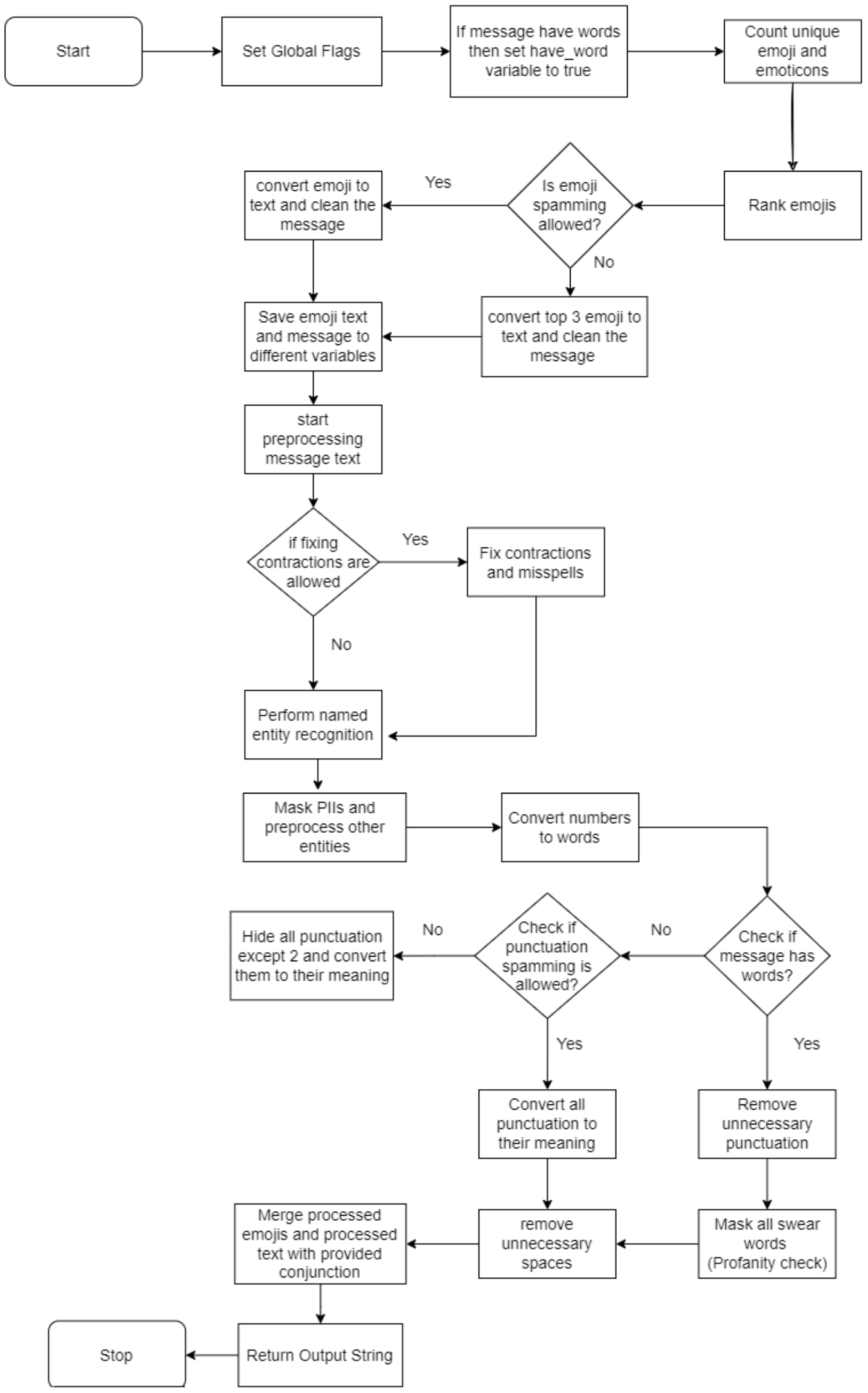

Fig. 1. Improved TTS flowchart

Our basic principle is to tell the system, how we humans comprehend any message or text. Our solution is fast and lightweight, and it handles a variety of real-world issues –

1) **Load global user flags**
   - ALLOW_PUNCTUATION_SPAMMING = False
   - ALLOW_EMOJI_SPAMMING = False
   - DISABLE_PII_MASKING = False
   - SHOW_PHONENUMBER = True

- RM_COMMON_ABBR = True

The first two flags control whether user want to listen streak of punctuation and emoji or not. DISABLE_PII_MASKING flag will disable the algorithm to mask PII. SHOW_PHONENUMBER flag allows to unmask phone numbers, keeping other PII masked and RM_COMMON_ABBR controls whether the algorithm corrects message contractions and misspellings.

2) **Convert message to lowercase**
3) **Using a regular expression, check whether the message contains any characters (A-Z, a-z, 0-9) and save the result in the have_char variable.**
4) **Count the total number of unique emojis in the message**
5) **Extract emoji from the message and rank them based on popularity-**
   a) Load *emoji_rank.json* file that has emoji meaning as key and its popularity.
      i) Eg - {*"rolling on the floor laughing"*: 5000}
      ii) If house and star emoji have 50 and 60 popularity respectively then star emoji will come first.
      The message - *"House and star emoji* 🏠⭐*"* is converted to - *"House and star emoji* ⭐🏠*"*
   b) Loop through each emoji in the message finds its meaning and its popularity and save it in a dictionary (rank_dict) like this - {*"emoji_meaning"*: popularity}
      i) We have removed the skin tone, for example 🧒🏼and 🧒means *"child"* emoji.
      ii) We have removed skin tone because it does not add any value while reading a message, for example - *He is a brave child* 🧒🏼🧒 is converted to *he is a brave child with child emoji.*
      iii) If emoji represents information, for example - 4️⃣ , then don't add it to rank_dict as we want to preserve their position.
         - For example - *"I will be there in* 4️⃣ *hour."*
      iv) Why we don't add 4️⃣ in rank_dict? Because all the emoji in rank_dict is added at the end of the message.
         A) Message - *"I will be there in* 4️⃣ 🤣🤣*hours."* is converted to - *"I will be there in four hours with rolling on the floor laughing emoji"*.
         B) But if we don't preserve their position then the output is - *"I will be there in hours with four and rolling on the floor laughing emoji"*.
         C) Emoji that can preserve their position - Number emoji, Alphabet emoji.
   c) Sort the rank_dict dictionary by value and concatenate the keys and save them in a variable (processed_emoji)
      i) After sorting - {*"star"*: 600, *"house"*: 500}
      ii) To prevent spamming concatenate top three emoji and hide others by *"and some other emoji"*.
      For example - *"star emoji, house emoji, child emoji and some other emoji"*.
      iii) If the message has less than 3 emoji, then after concatenating - *"star emoji, house emoji"*.
6) **Convert information-representing emoji to their meaning (Eg - '** 4️⃣ **'), their position are preserved and save the processed message to variable processed_text**
   - For example - *"I will be there in* 4️⃣ 🤣🤣*hours"*
      a) After extraction and ranking of emoji, the message is converted to - *"I will be there in* 4️⃣ *hours"*. As we have saved non-informational emoji in another variable, processed_emoji.
      b) And after converting ' 4️⃣ ', the message became - *"I will be there in four hours"*.
      c) Now save this message in a new variable, processed_text.
7) **If have_char is true then execute steps 8-11. All these operations are executed and output is saved in processed_text variable.**
8) **Fix message contractions and common misspell**.
   a) We have created a dictionary of commonly used contractions and misspells and their correct form.
   For example - {*"2day"*: *"today"*, *"msg"*: *"message"*}
   b) Text Normalisation -
      i) If users emphasize a word it may not appear in the dictionary; therefore, to address this issue, -
         A) Convert word to its unique form by using regular expression,
         For example - "message" or "messsaaageee" is converted to "mesage".
         B) Now "mesage" is converted to its correct form i.e. "message"
9) **Extract important entity from the message using regular expression.**
   We have extracted links, emails, percentages, hashtags, mentions, currency, length – (feet, inches), National ID, driving licence, phone numbers, time, date, decimals, and fractions. It is crucial that each word should be scanned by regular expression so that the date regex does not capture *"2018/10/03"* from link *"www.example.com/2018/10/03"*.
   Examples of a word is "$ 100" or "$100" or *"website www.example.com is ranked one in google"* have seven words *"website"*, *"www.example.com"*, *"is"*, *"ranked"*, *"one"*, *"in"*, *"google"*.
10) **Mask all personally identifiable information like phone number, national ID, driving license.**
11) **Convert all numbers to words as "2" to *"two"*.**
12) **Mask all swear words with *"beep"* using regular expression.**
   a) We have collected over a thousand swear words.
   b) Converted them as a regular expression like this - "word1|word2|word3".
   c) Replace all matches with *beep*.
13) **If the message has character (have_char is true) then remove all punctuation except question mark,**

**full stop, exclamation mark ,and comma as they are used in intonation otherwise convert the first three punctuation to their meaning and save it to processed_text variable.**
For example - *"?#@"* is converted to *"question mark hash symbol at symbol"*. If the message has more than three symbols then *"?#@&*()"* is converted to *"question mark hash symbol at symbol and some other punctuations"*.

14) **Concatenate processed_text and processed_emoji with the conjunction *"with"*.**
For example - *"I will be there in 4️⃣ 🤣🤣hours."*
is converted to : *"I will be there in four hours with rolling on the floor laughing emoji"*

## IV. Observations

We have selected runtime as KPI for our pre-processor which can parse a 50 character string in 4 ms while keeping message information remain intact. However, because emoji ranking is based on emoji popularity, emojis are not always read as expected, so we chose the longest contiguous matching subsequence as our KPI, which is 83.11%. Emojis were designed to express emotions through text, but when they are spoken like regular text, the entire meaning of the message is lost. **We refer to the existing Google TTS on Android 11 as E-TTS and our improved TTS as LIP-TTS.**

### *A. Handling message with emoji and text*

| Message | *The show was awesome 🤣🤣I liked it* |
|---|---|
| E-TTS | *The show was awesome rolling on the floor laughing rolling on the floor laughing I really liked it* |
| LIP-TTS | *the show was awesome i really liked it with rolling on the floor laughing emoji* |

### *B. Emoji Spamming*

We never read punctuation or emoji streaks, but we notice that messages include simply punctuation and discover unique emojis to comprehend the emotion. Our improved TTS system recognises emoji spamming and will only read popular emoji.

| Message | 🎁🍕✨😂🎉😎🤣 |
|---|---|
| E-TTS | *wrapped gift pizza sparkles face with tears of joy party popper smiling face with sunglasses partying face rolling on the floor laughing* |
| LIP-TTS | *Face with tears of joy emoji sparkles emoji rolling on the floor laughing emoji and some other emojis.* |

The ranking of emoji changes the ordering of emoji in the output. However, upon closer inspection, we see that emojis depicting emotions are retained in the output.

### *C. Punctuation Spamming –*

LIP-TTS can identify messages with only punctuation (i.e. no informational message) and provides better output.

| Message | *!@#$%^&* |
|---|---|
| E-TTS | *exclamation mark at symbol hash symbol dollar sign percentage symbol caret ampersand sign* |
| LIP-TTS | *exclamation mark at symbol hash symbol and some other punctuations* |

### *D. Emoji with Punctuation Spamming*

| Message | *!@#$%😂🤣😊🎉* |
|---|---|
| E-TTS | *exclamation mark at symbol hash symbol dollar sign percentage symbol caret ampersand sign asterisk face with tears of joy emoji rolling on the floor laughing emoji smiling face with smiling eyes emoji party popper emoji* |
| LIP-TTS | *exclamation mark at symbol hash symbol and some other punctuation face with tears of joy emoji rolling on the floor laughing emoji smiling face with smiling eyes emoji and some other emojis* |

### *E. Text with emoji and punctuation spamming*

It is common for texts to contain text with a lot of punctuation and emojis. Now that our pre-processor is aware that the message contains text, and we cannot remove punctuation at random. We must ensure that punctuations used for intonation, symbols with meaning such as *'$'* in *"$ 100"*, *'%'* in *"100 %"* are preserved, and for this, we have developed an algorithm that can recognise valuable entities and pre-process them independently so that no information is lost.

| Message | *Yesss!!!!! Its holiday today 🥳🥳* |
|---|---|
| E-TTS | *yes <pause> its holiday today partying face partying face* |
| LIP-TTS | *yes its holiday today with partying face emoji* |

We can preserve the intonation of exclamation mark by setting RM_ALL_PUNCS to false

| | |
|---|---|
| LIP-TTS | *yes! its holiday today with partying face emoji* |

### *F. Ensuring if emoji has information then it should preserve its position.*

| Message | *I will be there in 4️⃣ hours* |
|---|---|
| LIP-TTS | *i will be there in four hours* |

### *G. Emoji Ranking – Ensuring emotion never lost*

If we have a message with text and a lot of emojis, TTS will read only three emojis and the rest will be hidden, thus we may lose those emojis that convey the emotion of the message. To address this issue, we ranked the emojis based on their popularity.

| Message | *I have bought my own house 🏡🏡🏡🤩🤩🤩how serene this place is 🕺🕺🕺yeyyeye ✨✨* |
|---|---|
| E-TTS | *I have bought my own house house with garden house with garden house with garden star-struck star-struck star-struck how serene this place is man dancing man dancing man dancing yeyyeye sparkles sparkles* |
| LIP-TTS | *i have bought my own house how serene this place is yeyyeye with sparkles emoji star struck emoji man dancing emoji and some other emojis* |

### *H. Identification of PII (Personally Identifiable Information)*

We have developed a feature to identify and mask personally identifiable information (PII) such as emails, driving licence numbers,Aadhar card numbers (currently only for India), and phone numbers. We are expanding this library to include support for various nations.

| Message | *Email address - mohan@gmail.com, Phone number - 9321673878* |
|---|---|
| E-TTS | *email address mohan at gmail dot com phone number nine billion three hundred and twenty-one million six hundred and seventy-three thousand eight hundred and seventy-eight* |
| LIP-TTS | *email address email phone number a ten digit long number* |

### I. Improving how phone numbers are spoken in existing TTS

E-TTS translate phone number to words that are very inconvenient to hear. As a result, we upgraded this feature, and LIP-TTS reads it like we humans do.

| | |
|---|---|
| Message | *Phone number - 9321673878* |
| E-TTS | *Phone number nine billion three hundred and twenty-one million six hundred and seventy-three thousand eight hundred and seventy-eight* |
| LIP-TTS | *phone number nine three two one six seven three eight seven eight* |

### J. Improving how URLs are spoken in existing TTS

E-TTS reads a link from end to end, i.e. everything from protocol to domain name to everything in it. In real life, though, we generally care about domain names and, if possible, where page links will take us. For example, in *"https://cloud.google.com/natural-language"* it is easy to find that we are navigated to the natural language page on cloud.google.com, but if the link is too long, it is difficult to detect and we normally do not care. As a result, we will recognise links in the message and will only state *"domain name"* if we are on the homepage; otherwise, we will say *"webpage on domain name"*.

| | |
|---|---|
| Message | *homepage - https://www.google.com/ somewhere else - https://cloud.google.com/natural-language* |
| LIP-TTS | *homepage goggle dot com somewhere else webpage on cloud dot goggle dot com* |

### K. Profanity Check

Swear words are spoken as general text in the E-TTS, which can make users feel embarrassed if the system is active in public. So we have scraped numerous swear words from the internet to censor them with *"beep"*.

| | |
|---|---|
| Message | *Oh!!! Shit...what have u done?? fk off* |
| LIP-TTS | *oh beep what have you done beep off* |

### L. Text standardization and correction

In this fast-paced society, people use shortcuts that must be understood in order to be properly pronounced. Existing TTS is unable to pronounce it properly so we have scraped numerous shortcuts ($>$ 45k) used in day-to-day life which is used to fix this problem.

| | |
|---|---|
| Message | *I hv sent u a msg on fb, plz chk it n let me knw* |
| E-TTS | *I H V sent you a M S G on F B P L Z C H K it n let me K N W* |
| LIP-TTS | *i have sent you a message on Facebook please check it and let me know* |

Including this, we have also added functionality to fix common spell errors.

| | |
|---|---|
| Message | *I saw ur mssg on fbk* |
| E-TTS | *I saw ur M S G on F B K* (Note - *"ur"* is pronounced *"ur"* not *"your"*) |
| LIP-TTS | *i saw your message on facebook* |

### M. Entity Recognition

We have created a module to recognize entities from the messages so that their special meaning should be retained for example in *" 5' "*, the apostrophe has the meaning of feet but if we pre-process the message without considering this as length then we get erroneous results.

## V. Discussion and Insights

Our online survey[3] of 29 people of age group 21 to 55 years with a diversity of 31% and 69% of female and male respectively found:

1) All the surveyees (100%) liked the user-friendly way of parsing phone number and text normalization. For example - *"My phone number is 9321673878" to "My phone number is nine three two one six seven three eight seven eight" and "I hv sent u a msg on fb, plz chk it n let me knw" to "I have sent you a message on facebook please check it and let me know"*
2) Around 90% of them liked the way we pre-process messages with emojis. For example -*" I will be there in 4️⃣ 🤣🤣hours." is converted to "i will be there in four hours with rolling on the floor laughing emoji"*
3) Around 69% of them liked to mask PII, for example -*"This is my email address - mohan@gmail.com and Aadhar card number is 3675 9834 6012" is converted to "This is my email address email aadhar card number a twelve digit number"*. On asking the remaining 31% of them, why they don't want PII to be masked and selected *"This is my email address mohan at gmail dot com aadhar card number three thousand six hundred and seventy five nine thousand eight hundred and thirty four six thousand and twelve"* in the survey? They replied email should be unmasked.
4) Around 51% of them liked to rank emojis if there is a streak of emojis in the message. *F*or example - "🌟🎂🎈🙏☁️🎁🍕✨😂🎉😎🤣"
5) Around 48% of them liked to avoid reading streak of punctuation, for example - "!@#$%&*()".
6) Around 31% have selected to censor swear words.

Since all submodules are independent so they can be personalized by global flags and email can be unmasked keeping all other PII masked.

## VI. Error Estimation

Our proposed system works on almost every real-world scenario but there are cases when it may fail -

1) If there is any random 10 digit number, for example- *"Your username is 1234010120, with first five-digit is last four digits of your phone number and then your date of birth in DDMMYY format"*In this example "1234010120" is converted to *"one two three four zero one zero one two zero" but the expected output is "a ten-digit number"* as it is a PII.

[3]Survey results have been presented in Appendix.

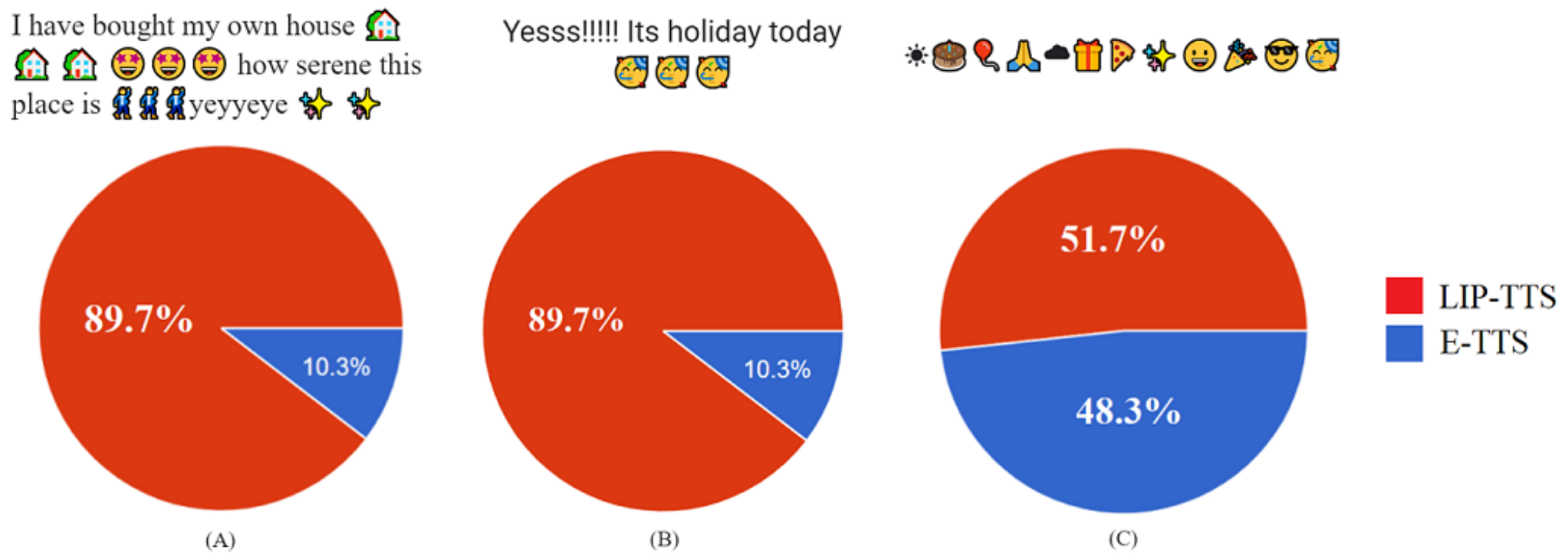

Fig. 2. Sample outcome of black-box questionnaire survey : 76.5% of participants preferred the output of LIP-TTS over E-TTS.

2) If emoji is used as an adjective, for example - *"Its a ☀️ day"* is converted to *"Its a day with sun emoji"* but the expected output is *"its a sunny day"*

## VII. Conclusion

We propose a lightweight intelligent preprocessor (LIP) that accepts a text input and converts it into a more TTS-friendly text output. This work is the first of its kind, primarily as it targets an end-to-end on-device inference on resource-constrained edge devices with the aim to enable an enhanced TTS output for accessibility. With this goal, our proposal has a memory footprint of only 3.55 MB. Furthermore, an inference time of 4 ms for up to 50-character text facilitates real-time performance on smartphones. In addition, our pipeline is designed to be modular and plug-and-play, making it customizable by end-user and potent for seamless future extension or language expansion for code-mixed scripts. LIP addresses the drawbacks of existing TTS systems while handling some popular content of text messages like spanned emojis, punctuation spamming, PIIs, URLs, profanity, SMS shorthands, etc. We benchmark with a black-box survey from end-users who evaluated outputs from existing TTS engines with and without our preprocessor in the pipeline. The analysis of the survey reports indicates an overwhelming 76.5% preference towards LIP-enabled TTS over the base TTS engine.

## VIII. Future Work

In order to minimise the on-device inference time, some of the sub-modules of LIP relies on preloaded static dictionaries, and regular expressions. Some of these like emoji ranking have scope to benefit from personalization based on user's own past usage of emoji subsets. Furthermore, on-device emphasis detection [14] and sentiment extraction [15] can be used to augment LIP with a specific TTS voice, and in effect, introduce appropriate intonation and nuances to the TTS output. We are also exploring code-mixed languages like Hinglish (Romanized Hindi), leveraging on-device language detection [16] to select language-specific models in the LIP pipeline.

## Appendix

The results of the online survey discussed in section V are presented as follows.

## Survey Results

Legends -
- LIP-TTS
- E-TTS

1. **Gender**

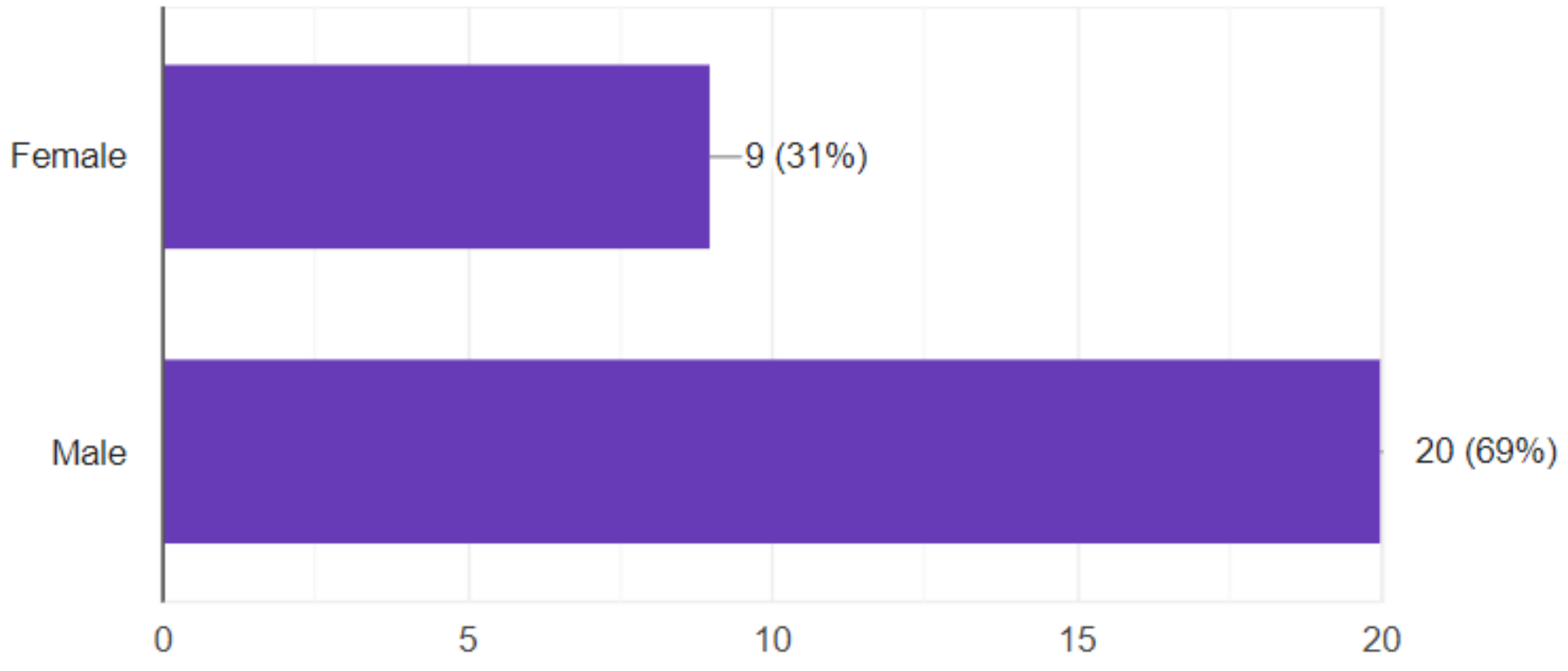

2. **I hv sent u a msg on fb, plz chk it n let me knw**

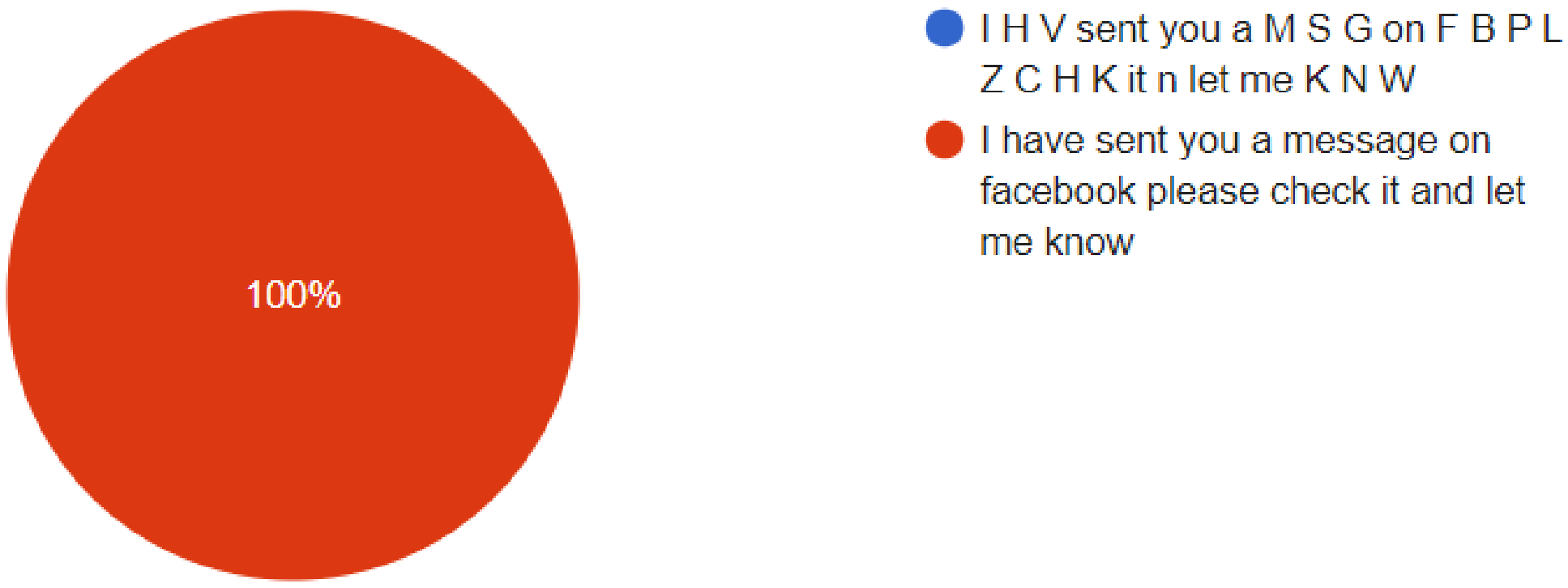

3. **My phone number is 9321673878**

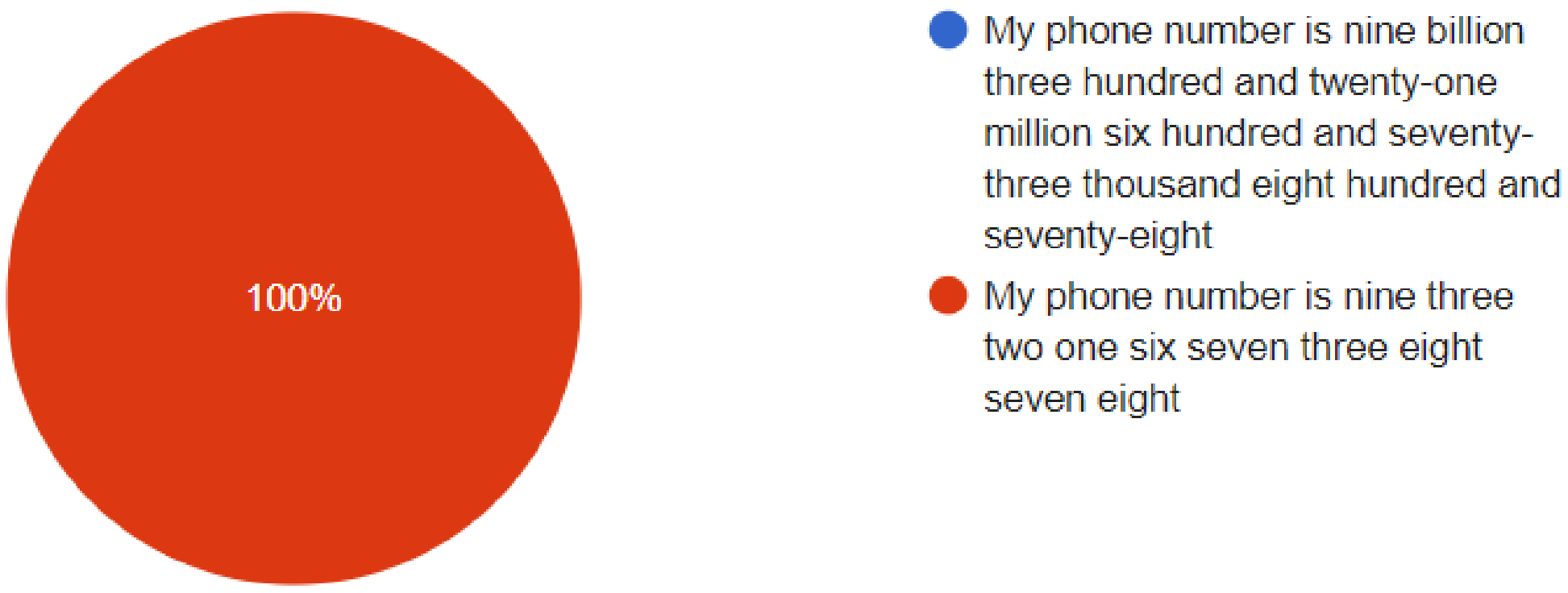

4. **Yesss!!!!! It's holiday today 🥳🥳🥳**

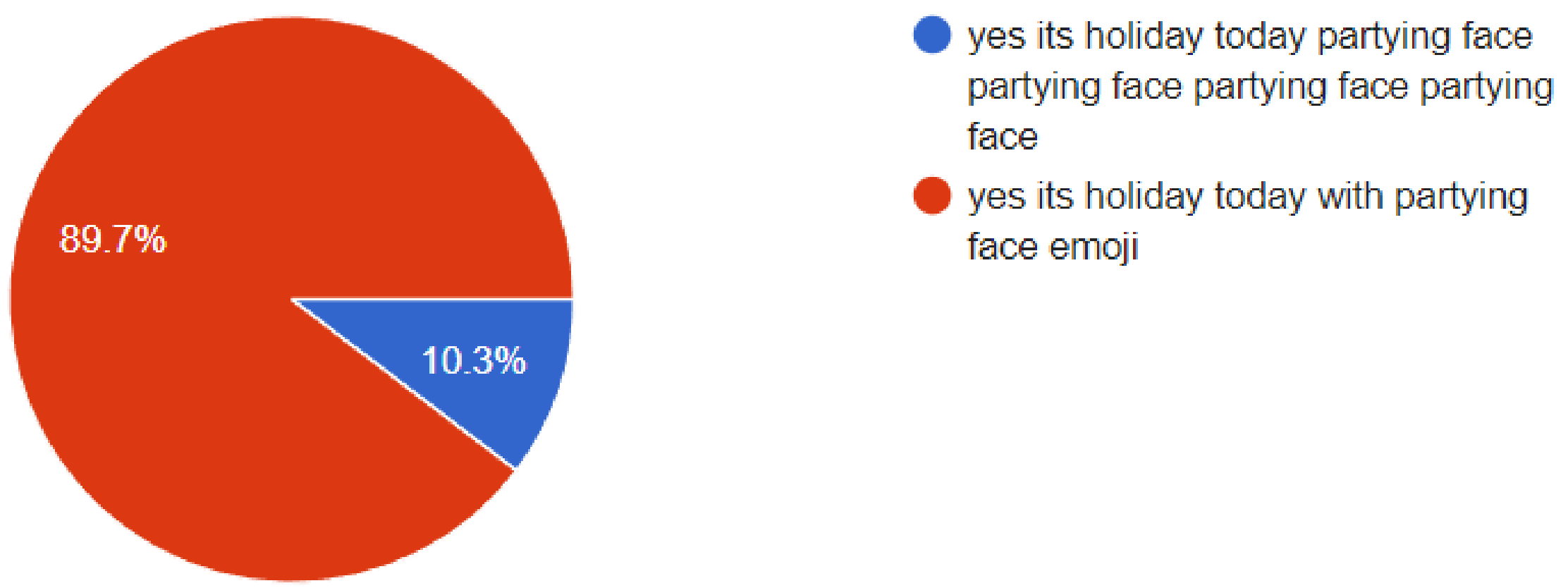

5. **I will be there in 4️⃣🤣🤣 hours.**

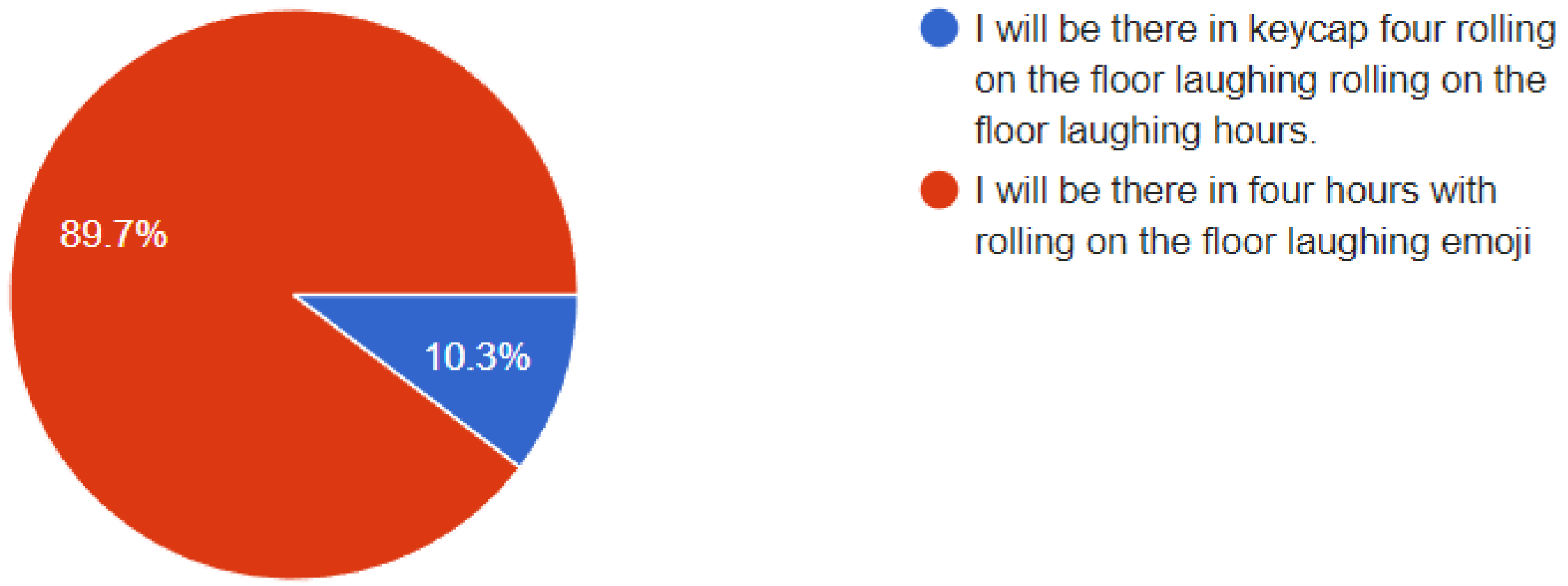

6. **I have bought my own house 🏡 🏡 🏡 🤩🤩🤩 how serene this place is 🕺 🕺 🕺 yeyyeye ✨ ✨**

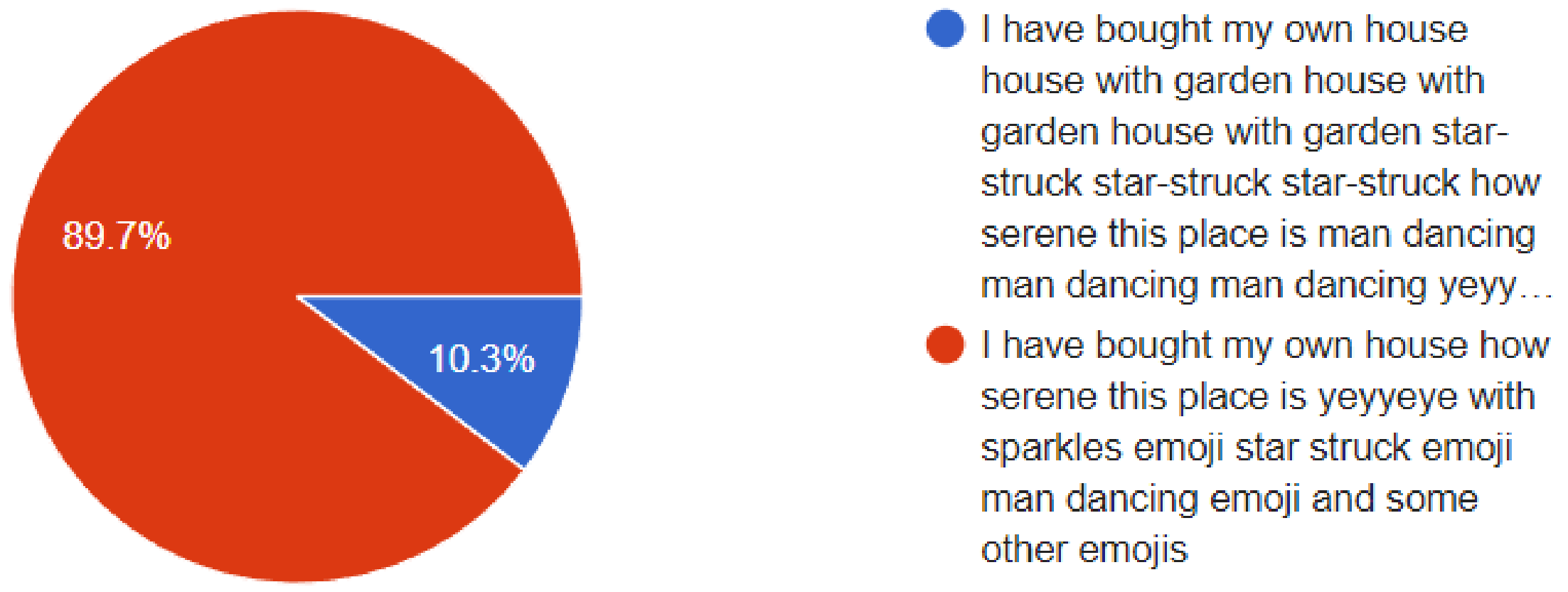

7. **The show was awesome 🤣🤣🤣 I liked it**

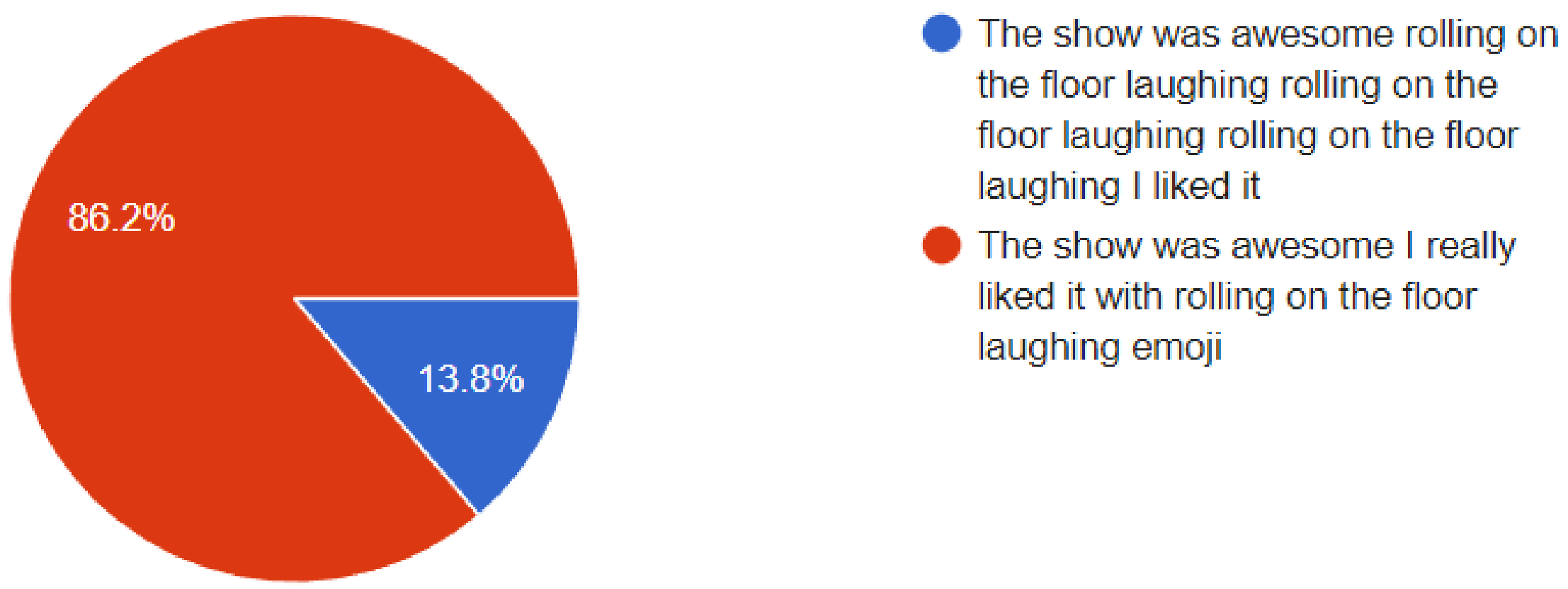

8. **!@#$%^&*🤣😂😊😁🎉😀🥳😎** *(Punctuation Spamming + Emoji Spamming)*

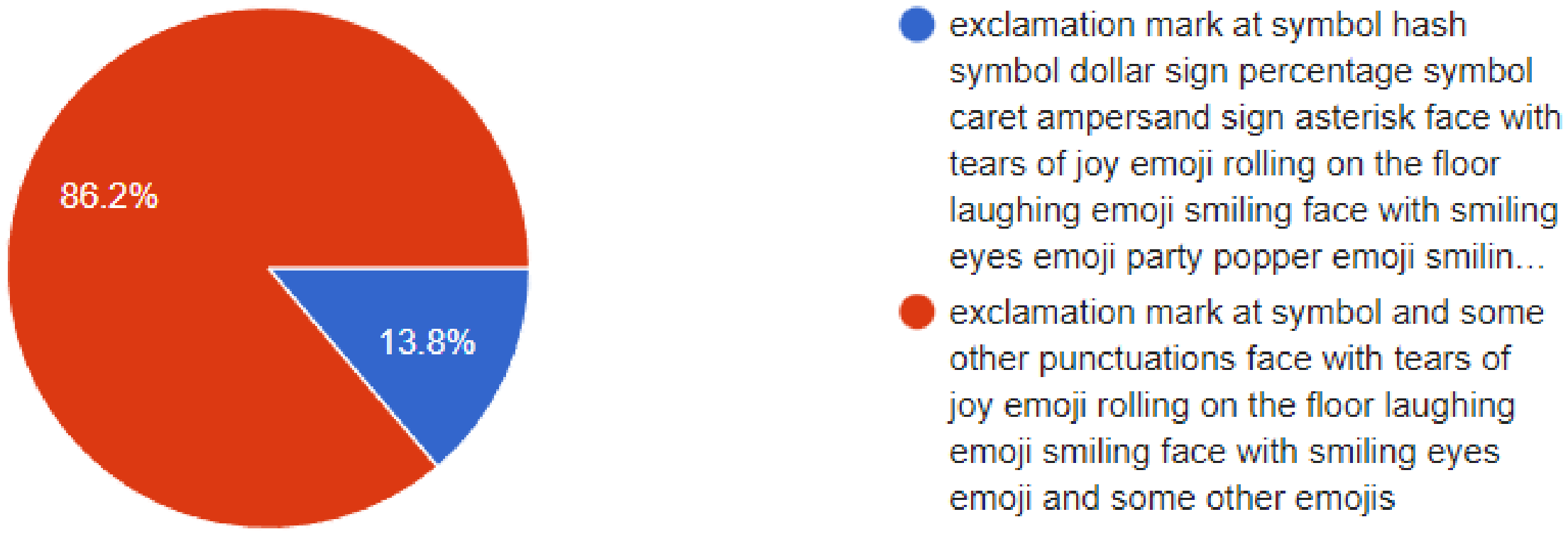

9. **This is my email address - mohan@gmail.com and Aadhar card number is 3675 9834 6012**

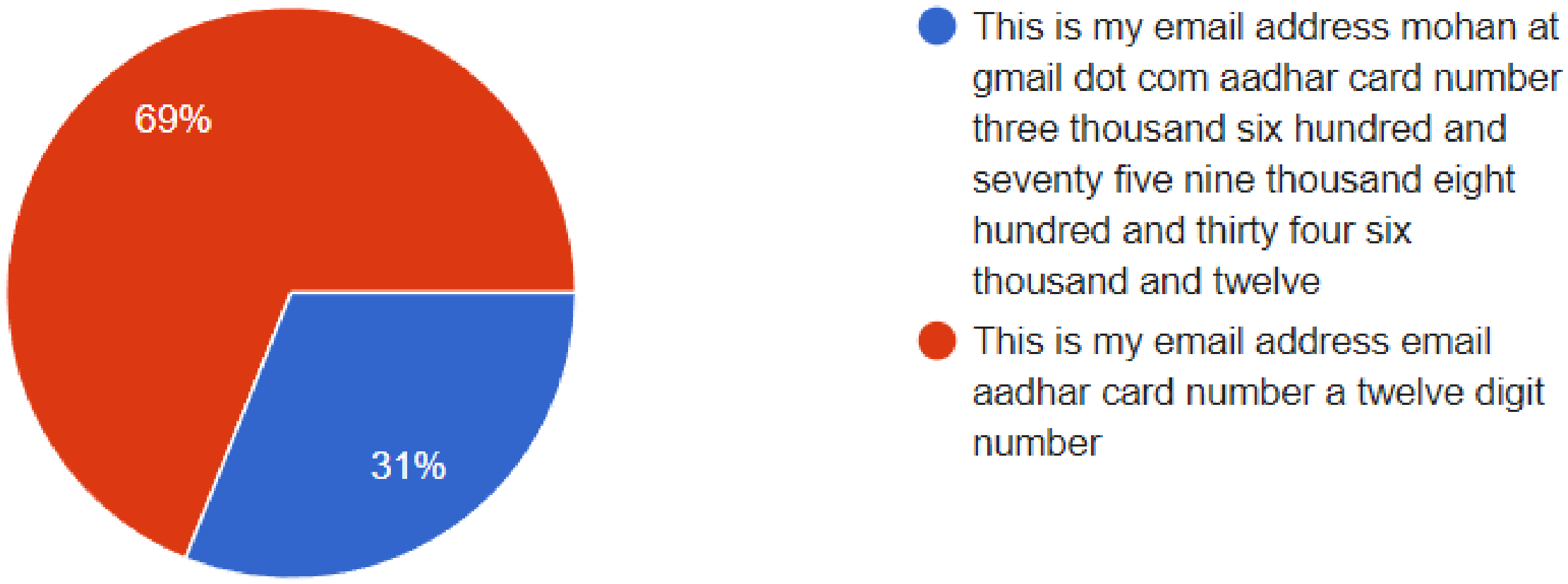

**10.☀🎂🎈🙏☁️🎁🍕✨😀🎉😎🥳** ***(Emoji Spamming)***

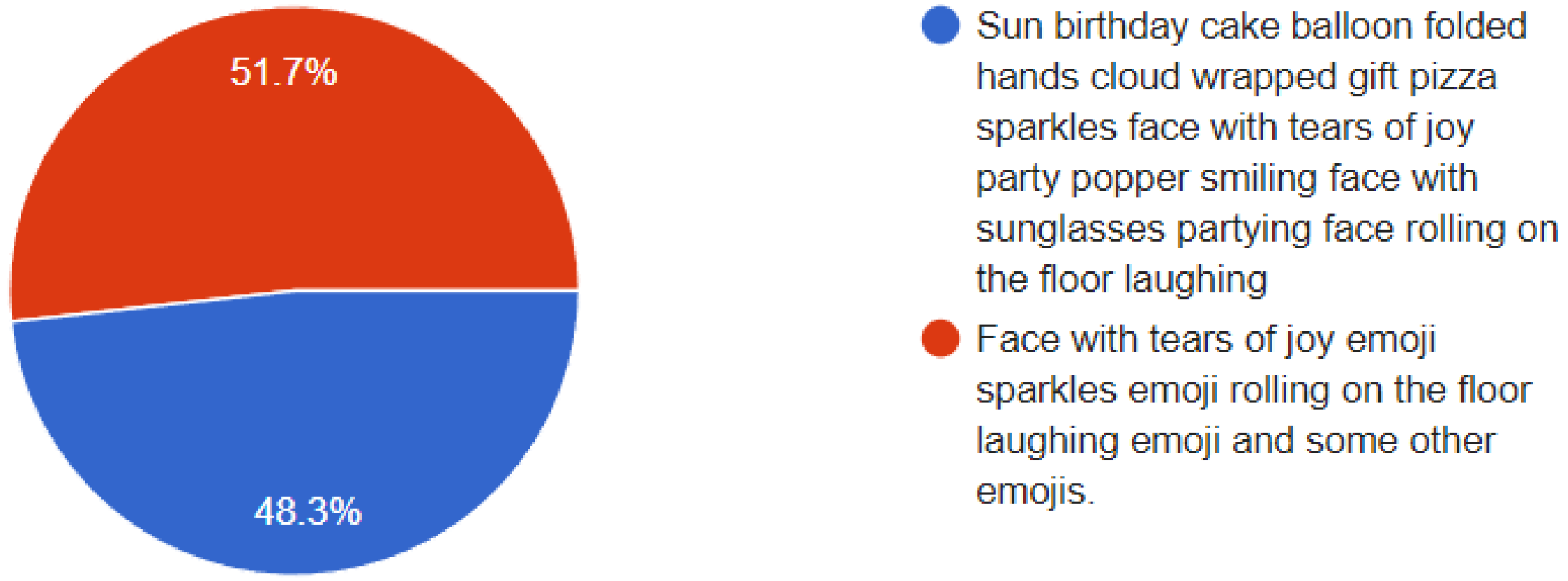

**11. !@#$%&*()** ***(Punctuation Spamming)***

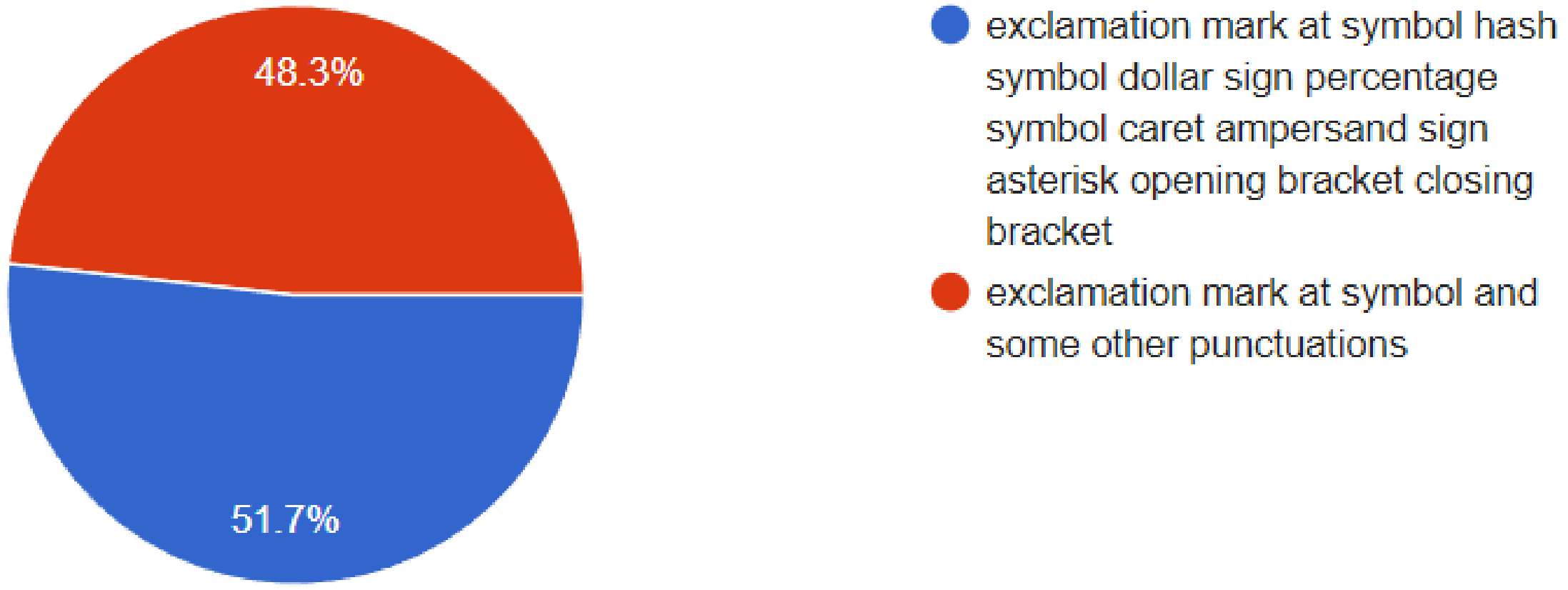

**12.Oh!!! Shit...I missed that question??**

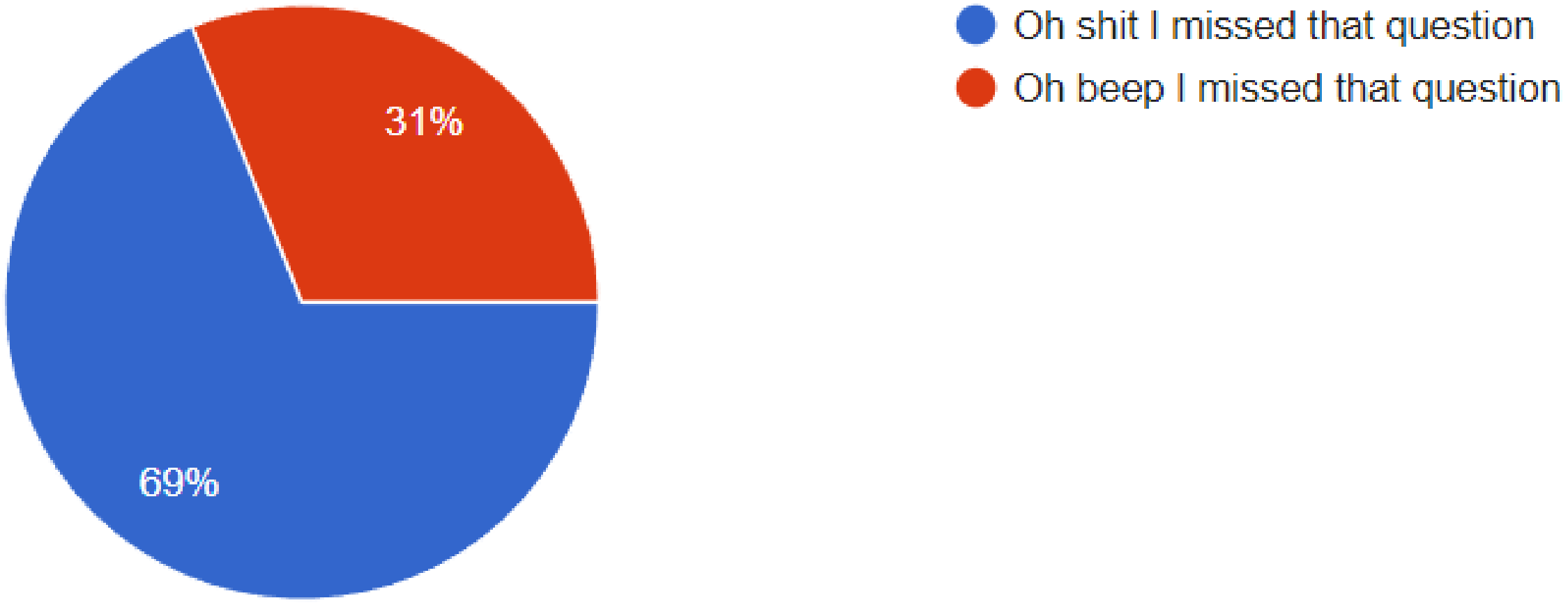